\documentclass[runningheads]{llncs}

\usepackage[mobile]{eccv}

\usepackage{eccvabbrv}

\usepackage{booktabs}
\usepackage{colortbl}
\usepackage{pifont}
\newcommand{\cmark}{\ding{51}}% checkmark
\newcommand{\xmark}{\ding{55}}% x mark
\usepackage{amsmath,amsfonts,bm}

\def\eqref#1{equation~\ref{#1}}
\def\1{\bm{1}}

\DeclareMathAlphabet{\mathsfit}{\encodingdefault}{\sfdefault}{m}{sl}
\SetMathAlphabet{\mathsfit}{bold}{\encodingdefault}{\sfdefault}{bx}{n}

\DeclareMathOperator*{\argmin}{arg\,min}

\usepackage{multirow}
\usepackage{colortbl}
\usepackage{todonotes}
\usepackage{algorithm}
\usepackage{algpseudocode}
\usepackage{pifont}
\definecolor{cvprblue}{rgb}{0.21,0.49,0.74}
\usepackage{listings}
\usepackage{xcolor}

\definecolor{codegreen}{rgb}{0,0.6,0}
\definecolor{codegray}{rgb}{0.5,0.5,0.5}
\definecolor{codepurple}{rgb}{0.58,0,0.82}
\definecolor{backcolour}{rgb}{0.95,0.95,0.92}

\lstdefinestyle{mystyle}{
    backgroundcolor=\color{backcolour},
    commentstyle=\color{codegreen},
    keywordstyle=\color{magenta},
    numberstyle=\tiny\color{codegray},
    stringstyle=\color{codepurple},
    basicstyle=\ttfamily\footnotesize,
    breakatwhitespace=false,
    breaklines=true,
    captionpos=b,
    keepspaces=true,
    numbers=left,
    numbersep=5pt,
    showspaces=false,
    showstringspaces=false,
    showtabs=false,
    tabsize=2
}

\usepackage[accsupp]{axessibility}  % Improves PDF readability for those with disabilities.
\usepackage{graphicx}
\usepackage{tikz}
\usepackage{comment}
\usepackage{amsmath,amssymb} % define this before the line numbering.
\usepackage{color}

\usepackage{booktabs}
\usepackage{multirow}
\usepackage{colortbl}
\usepackage{xcolor}
\usepackage{graphicx}
\usepackage{gensymb}
 \usepackage{adjustbox}
\usepackage{wrapfig}
\usepackage{caption}
\usepackage{comment}
\usepackage{cancel}
\usepackage{rotating} %
\usepackage{pifont}

\usepackage{multirow}

\usepackage[colorlinks,linkcolor=blue,urlcolor=blue]{hyperref}
\usepackage{orcidlink}

\begin{document}

% ---------------------------------------------------------------
% TODO REVIEW: Replace with your title
\title{Free-Editor: Zero-shot Text-driven 3D Scene Editing} 

% TODO REVIEW: If the paper title is too long for the running head, you can set
% an abbreviated paper title here. If not, comment out.
\titlerunning{Free-Editor}

% TODO FINAL: Replace with your author list. 
% Include the authors' OCRID for the camera-ready version, if at all possible.
\author{Nazmul Karim\inst{*,1}\orcidlink{0000-0001-5522-4456} \and Hasan Iqbal\inst{*,2}\orcidlink{0009-0005-2162-3367} \and
Umar Khalid\inst{*,1}\orcidlink{0000-0002-3357-9720} \and
Chen Chen\inst{1}\orcidlink{0000-0003-3957-7061} \and Jing Hua\inst{2}\orcidlink{0000-0002-3981-2933}
}

% TODO FINAL: Replace with an abbreviated list of authors.
\authorrunning{N.~Karim et al.}
% First names are abbreviated in the running head.
% If there are more than two authors, 'et al.' is used.

\institute{University of Central Florida, Orlando, FL, USA \and
Department of Computer Science, Wayne State University, Detroit, MI, USA}

\maketitle

{ \renewcommand{\thefootnote}%
    {\fnsymbol{footnote}}
% \footnotetext[1]{{\tt\scriptsize Email: nazmul.karim18@knights.ucf.edu}}
% \footnotetext[2]{{\tt\scriptsize Email: firstname.lastname@sri.com}}
  \footnotetext[1]{* Equal Contribution}
}

\begin{abstract}
% In our work, we aim to alleviate these issues by designing an intuitive technique that possesses the capability of editing 3D scenes without re-training the model.

Text-to-Image (T2I) diffusion models have recently gained traction for their versatility and user-friendliness in 2D content generation and editing. However, training a diffusion model specifically for 3D scene editing is challenging due to the scarcity of large-scale datasets. Currently, editing 3D scenes necessitates either retraining the model to accommodate various 3D edits or developing specific methods tailored to each unique editing type. Moreover, state-of-the-art (SOTA) techniques require multiple synchronized edited images from the same scene to enable effective scene editing. Given the current limitations of T2I models, achieving consistent editing effects across multiple images remains difficult, leading to multi-view inconsistency in editing. This inconsistency undermines the performance of 3D scene editing \footnote{Here, 3D scene editing indicates NeRF model editing. In this study, we mainly focus on NeRF-based 3D scene representation.} when these images are utilized. In this study, we introduce a novel, training-free 3D scene editing technique called \textsc{Free-Editor}, which enables users to edit 3D scenes without the need for model retraining during the testing phase. Our method effectively addresses the issue of multi-view style inconsistency found in state-of-the-art (SOTA) methods through the implementation of a single-view editing scheme. Specifically, we demonstrate that editing a particular 3D scene can be achieved by modifying only a single view. To facilitate this, we present an Edit Transformer that ensures intra-view consistency and inter-view style transfer using self-view and cross-view attention mechanisms, respectively. By eliminating the need for model retraining and multi-view editing, our approach significantly reduces editing time and memory resource requirements, achieving runtimes approximately 20 times faster than SOTA methods. We have performed extensive experiments on various benchmark datasets, showcasing the diverse editing capabilities of our proposed technique. Project Page: \url{https://free-editor.github.io/}
\end{abstract}    
\section{Introduction}
\label{sec:intro}
%\vspace{-1mm}
Neural Radiance Fields (NeRF)~\cite{mildenhall2021nerf}, neural implicits~\cite{wang2021neus} as well as subsequent work~\cite{liu2020neural, muller2022instant, wang2022nerf}, collectively termed as \textit{neural fields}, have emerged as powerful 3D neural representations. Recent advances in this field~\cite{fang2023one, fang2023pvd} have focused on both novel view synthesis, scene reconstruction as well as 3D scene manipulations such as color editing~\cite{kobayashi2022decomposing, wang2022clip,karim2023save,khalid2023latenteditor,zhuang2023dreameditor}, scene composition~\cite{tancik2022block, tang2022compressible}, and style transfer~\cite{gu2021stylenerf, hollein2022stylemesh}. Notably, it has been shown that text-guided 3D NeRF~\cite{zhuang2023dreameditor, haque2023instruct} editing can be achieved through leveraging the diverse generation capability of 2D text-to-image (T2I) diffusion models~\cite{brooks2023instructpix2pix, ruiz2023dreambooth, han2023svdiff}. Despite their \begin{wrapfigure}{r}{0.425\textwidth} 
%\begin{center}
\centering
% %\vspace{-3mm}
%\begin{minipage}{0.35\textwidth}
    \includegraphics[width=0.9\linewidth]{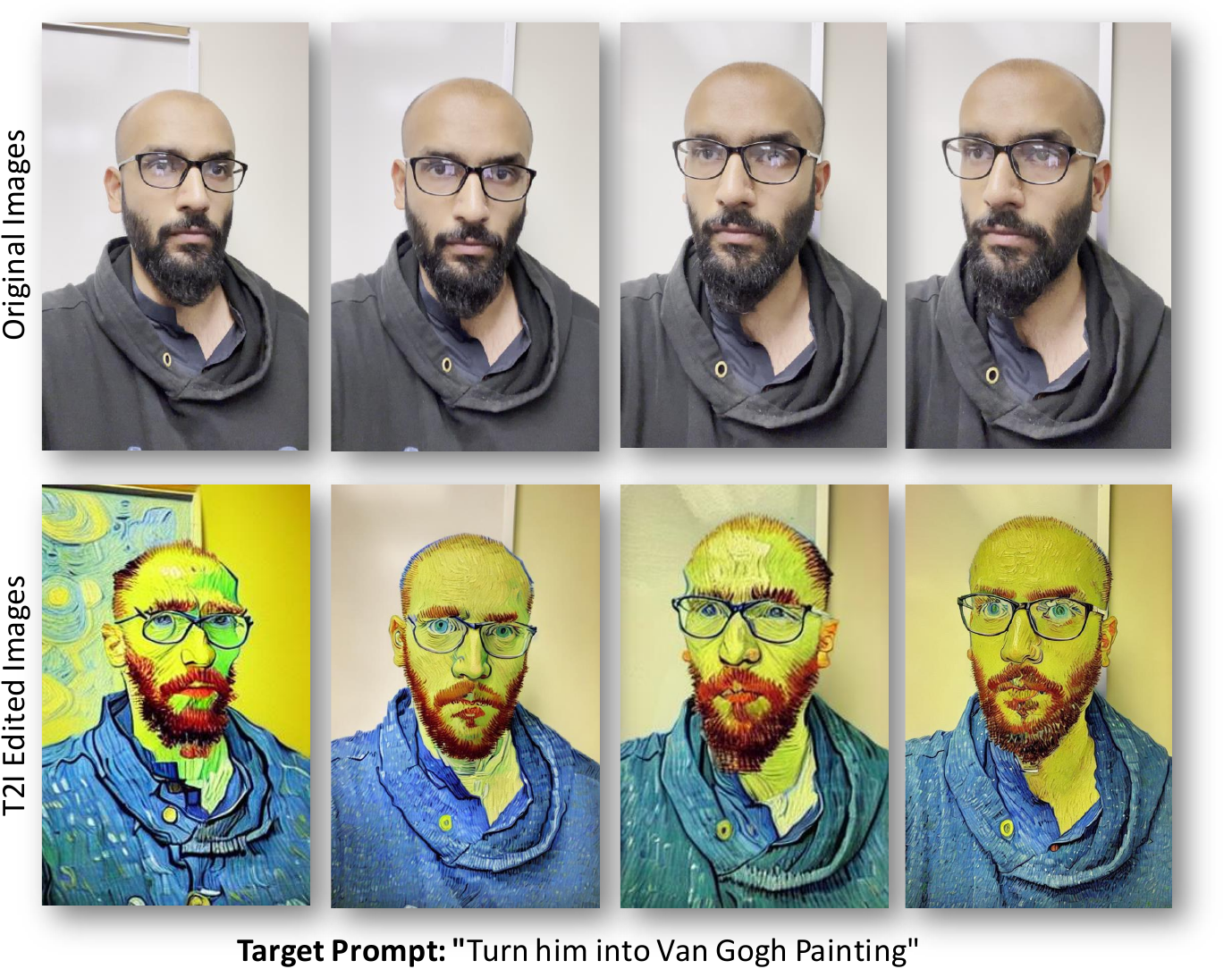}
%    \end{center}
    % %\vspace{-1.5mm}
    \caption{\footnotesize \textbf{Multi-View Inconsistency in Current Text-to-Image (T2I) Editing Models}: The current T2I editing model~\cite{brooks2023instructpix2pix} faces significant challenges with multi-view consistency. This issue adversely affects the quality of 3D scene editing, especially when these edited views are used to synthesize novel views. This specific limitation is also acknowledged in IN2N~\cite{haque2023instruct}. Note that this inconsistency is particularly problematic when editing is performed \emph{without re-training}, which aligns with our objectives.}
\label{fig:res_multi}
% %\vspace{-6mm}
\end{wrapfigure}demonstrated success existing methods need to \textbf{i)} re-train the editing model for each particular 3D scene which introduces computational and memory overhead, and \textbf{ii)} rely on the prior knowledge of specific editing types, which may not be feasible in most scenarios. For instance,  InstructNerf2Nerf (IN2N)~\cite{haque2023instruct} iteratively edits the training images of a scene until it obtains the desired editing result of the scene. 
The iterative editing is inevitable since all training images may not have consistent style information during initial iterations. This is due to the current limitations of T2I diffusion models as achieving prompt-consistent edits in multiple images (even if they are from the same scene) is very challenging. Fig.~\ref{fig:res_multi} shows an example where the same target prompt produces different multi-view outcomes within the scene. Such inconsistent edits or styles lead to poor 3D editing performance even with extensive re-training. To tackle this issue of \emph{multi-view style inconsistency}, IN2N proposes to iteratively update the edited training set based on direct feedback from the NeRF model. Achieving the same goal as IN2N without re-training limits us from updating the training set more than once. This leads to poor performance due to the aforementioned issues.

In our work, we tackle the problem of \emph{text-driven 3D scene editing} from a  fresh perspective. Given a 3D scene data with multiple source views with their pose information, we randomly choose a \emph{starting view}. Our objective is to edit the entire 3D scene by editing only this starting view. By editing only one view per scene, we eliminate the possibility of multi-view style inconsistency (Fig.~\ref{fig:res_multi}) while reducing the overall editing time significantly. In addition, we address the problem of re-training with the help of a generalized NeRF (Gen-NeRF) model~\cite{sajjadi2022scene,wang2022attention}. Specifically, we leverage the style information in the starting view and multi-view geometry information from several unedited source views to render a novel edited target view with the same style as the starting view. We argue that zero-shot 3D scene editing can easily be achieved by introducing a few key architectural design changes to the Gen-NeRF. 

\begin{table}[t]
\begin{minipage}{0.445\textwidth}
\centering
\caption{\footnotesize \textbf{Comparison with SOTA}. Unlike prior works, our proposed method does not re-train the model each time we have to edit a new scene.}  
    \scalebox{0.6}{
    \begin{tabular}{lccc}
    \toprule
%\multirow{2}[1]{*}
{Methods}   
% &  \multicolumn{1}{c}{Editing Requirements} &    \multicolumn{2}{c}{Editing Capacity}  \\
% \cmidrule{2-3} \cmidrule{4-5}  
&   Re-Training     & Text-Driven  & Style Transfer  \\
    \midrule
    \textbf{Blend-NeRF}~\cite{kim20233d} & \cellcolor{blue!25}\cmark& \cellcolor{red!25}\xmark &  \cellcolor{red!25}\xmark   \\
    \textbf{Blended-NeRF}~\cite{gordon2023blended} & \cellcolor{blue!25}\cmark & \cellcolor{blue!25}\cmark& \cellcolor{blue!25}\cmark\\
    \textbf{DreamEditor}~\cite{zhuang2023dreameditor} &\cellcolor{blue!25}\cmark& \cellcolor{blue!25}\cmark& \cellcolor{blue!25}\cmark\\
    
    \textbf{NeRF-Art}~\cite{wang2023nerf} & \cellcolor{blue!25}\cmark& \cellcolor{blue!25}\cmark& \cellcolor{blue!25}\cmark  \\
    \textbf{Instruct-N2N}~\cite{haque2023instruct} & \cellcolor{blue!25}\cmark& \cellcolor{blue!25}\cmark& \cellcolor{blue!25}\cmark  \\
    \textbf{Ours} & \cellcolor{red!25}\xmark& \cellcolor{blue!25}\cmark   &\cellcolor{blue!25}\cmark  \\
    \bottomrule
    \label{tab:comparison}
    \end{tabular}}%
\end{minipage}
\hfill
\begin{minipage}{0.475\textwidth}
    \centering
%\hspace{-5mm}
%\begin{minipage}{0.35\textwidth}
    \includegraphics[width=1\linewidth]{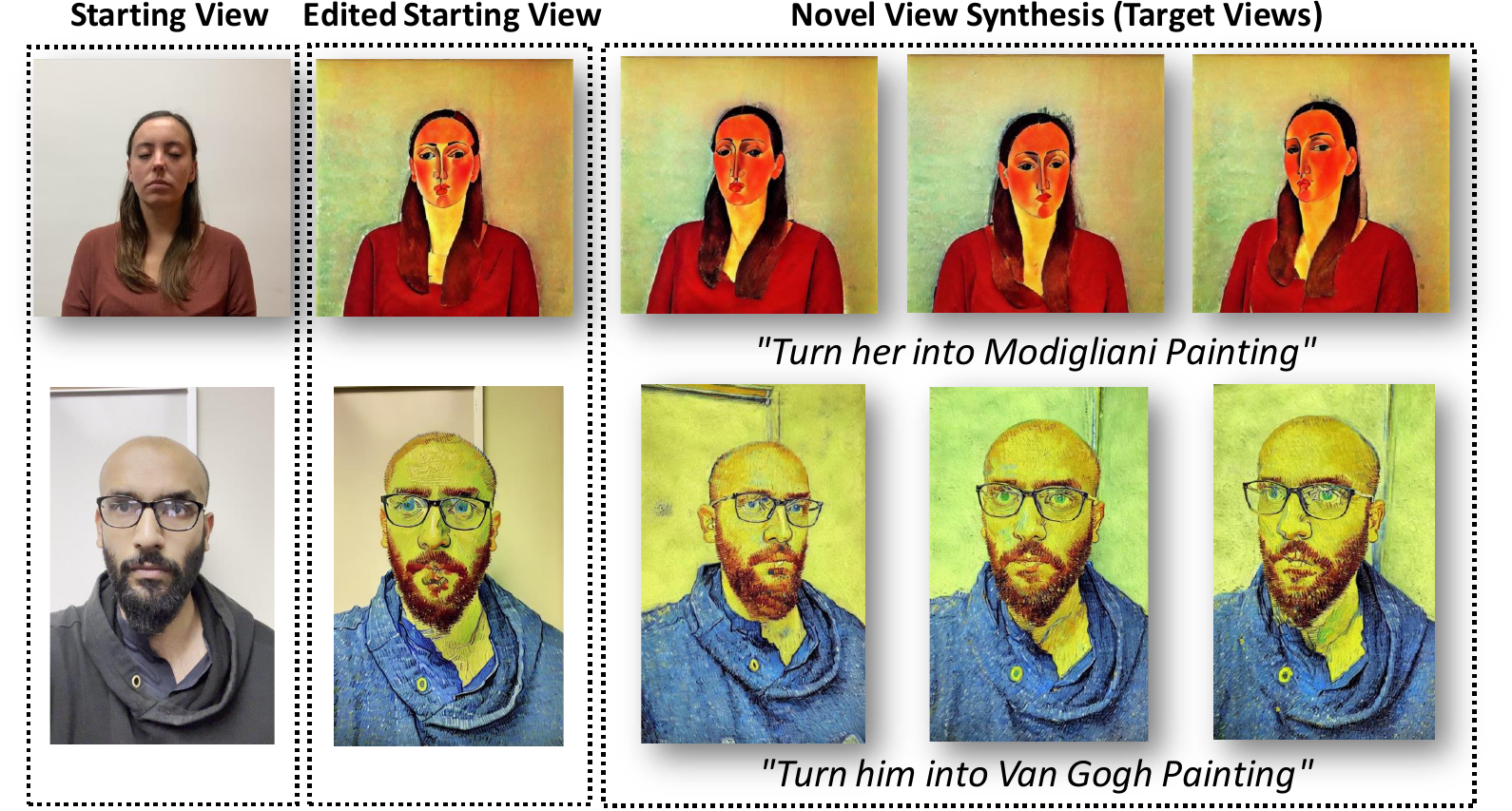}
%    \end{center}
    %\vspace{-4mm}
    \captionof{figure}{\footnotesize \textbf{3D Scene Editing} using our proposed method for different target poses.}
    % \caption{ }
    %\vspace{-7mm}
\label{fig:teaser}
\end{minipage}
    % %\vspace{-5mm}
\end{table}

These architectural changes are necessary as there are a few obvious challenges in achieving our goal: \textit{First}, transferring style information from the starting view to the target view requires the view geometry correspondence information. In our framework, the view-geometry correspondence is solved by leveraging pixel-aligned features for each target pixel. To further enhance these features (obtained from unedited source views) with style information, we utilize an \emph{Edit Transformer (ET)} that employs both self-view and cross-view attentions. While self-view attention helps us grasp long-range content information within the starting view, we can enrich the pixel-aligned features with content details from the starting view with cross-view attention. Subsequently, these style-informed features, obtained from ET, can easily be converted into RGB color using widely used Epipolar and Ray transformers~\cite{yang2022mvs2d, suhail2022generalizable}. \textit{Second}, features obtained from ET lack necessary spatial awareness as closely situated neighboring views from the same scene should change continuously. This may lead to spatial non-smoothness in the pixel space which is highly undesirable for style transfer. To tackle this, we design a multi-view consistency loss that encourages the features corresponding to two spatially close points to be similar. In addition, we employ \emph{self-view robust loss} to obtain consistent color in the final edited scene. Our main contributions can be summarized as follows: 
\begin{itemize}
    \item  We propose $\textsc{Free-Editor}$, a zero-shot text-guided 3D scene editing technique that can synthesize edited novel views based on a text description while maintaining high 3D consistency. By introducing several novel modifications to a generalized NeRF model, our proposed method eliminates the requirement of scene-specific retraining across various editing styles.
    \item  We propose an \emph{Edit Transformer} to facilitate intra-view consistency and inter-view style transfer, enabling us to edit a particular 3D scene using a single edited view (see Fig.~\ref{fig:teaser}). This \emph{single-view editing scheme} can effectively remove the bottleneck of \emph{multi-view style inconsistency} in SOTA methods.  
    \item Due to the unique design choices in $\textsc{Free-Editor}$, both training costs as well as editing time are reduced significantly. Extensive evaluation on multiple datasets with various editing styles demonstrates the superiority of our proposed method.
\end{itemize}

\section{Related Work}
%%\vspace{-1mm}
\noindent \textbf{Novel View Synthesis With NeRF.} was first introduced in~\cite{mildenhall2021nerf}, generates realistic novel views by fitting scenes as continuous 5D radiance fields using a Multilayer Perceptron (MLP). Since its inception, several advancements have enhanced NeRFs. For instance, Mip-NeRF~\cite{barron2021mip, barron2022mip} efficiently handles object scale in unbounded scenes. Nex~\cite{wizadwongsa2021nex} models significant view-dependent effects whereas other works~\cite{oechsle2021unisurf, wang2021neus} improve surface representation, extend to dynamic scenes~\cite{park2021nerfies}, etc. Despite the tremendous success of NeRF, its time-consuming per-scene fitting poses a notable drawback. To address this, Generalizable NeRFs aim to bypass this optimization hurdle by framing new view creation as an image-based interpolation challenge across different views. Approaches like NeuRay~\cite{liu2022neural}, IBRNet~\cite{wang2021ibrnet}, MVSNeRF~\cite{chen2021mvsnerf}, and PixelNeRF~\cite{yu2021pixelnerf} construct a universal 3D representation using combined features from observed views. GPNR~\cite{suhail2022generalizable} and GNT~\cite{wang2022attention} elevate the quality of generated new views through a Transformer-based aggregation method. In our work, we also consider Generalized NeRF as we do not aim to re-train the model.

\noindent \textbf{Diffusion-based 3D Scene Editing.} The emergence of text-to-image conversion models has notably impacted NeRF editing. Beginning with the Score Distillation Sampling method in DreamFusion~\cite{poole2023dreamfusion}, Vox-e~\cite{sella2023vox} explored techniques to regulate alterations in pre-existing voxel fields. NeRF-Art~\cite{wang2023nerf} utilizes various regularization approaches during training to ensure that NeRF when edited using CLIP, preserves the original structure. InstructNerf2Nerf (IN2N)  employed 2D image translation models~\cite{haque2023instruct} to adjust 2D image characteristics for NeRF training based on textual prompts. However, IN2N's reliance on IP2P~\cite{brooks2023instructpix2pix}  for updating NeRF training data tends to excessively modify scenes. Furthermore, encounter difficulties such as extended training durations and unstable loss functions. Addressing the challenge of undesired alterations, D-Editor~\cite{zhuang2023dreameditor} introduced a mesh-based neural field that efficiently converts 2D masks into 3D editing areas. This enables precise local modifications while avoiding unnecessary geometric changes when altering only the appearance. Similarly, Blended-NeRF~\cite{gordon2023blended} and Blend-NeRF~\cite{kim20233d} require additional cues like bounding boxes for localized editing. However, all of these methods require per-scene adaptation of the 3D model to induce any editing effects which increases computational overhead significantly. In our work, we propose a zero-shot editing technique that performs similarly or better than SOTA with more practical applicability.     
%\vspace{-3mm}
\section{Method}
%\vspace{-1mm}
\subsection{Background}
%\vspace{-1mm}
\textbf{Neural Radiance Fields.} In neural radiance fields (NeRF)~\cite{mildenhall2021nerf}, the task is to find a neural network-based representation of 3D scenes. The neural network here is a multi-layer perceptron (MLP) that maps a 3D location $\bm{x} \in \mathbb{R} ^{3} $ and viewing direction $\bm{d} \in \mathbb{S} ^{2} $ to {an} emitted color $\bm c \in \left [ 0,1 \right ]^{3}$ and a volume density $\sigma \in \left [ 0,\infty \right )$, 
\begin{equation}\label{eq:nerf}
    \mathcal{F}(\bm x, \bm d ; \bm \Theta) \ \ \mapsto \ \  \left ( \bm c, \sigma \right ),
\end{equation}
where $\mathcal{F}$ and $\bm \Theta$  represent MLPs and the set of learnable parameters, respectively.

\noindent \textbf{Volume Rendering.} Let us define $\bm r \left ( t \right)= \bm{o}+{t}\bm{d}$ as a ray in a NeRF, where $\bm{o}$ is the camera center and $\bm{d}$ is the ray's unit direction vector. Along this ray, we can predict the color values $\bm c_{i}$ and volume densities $\sigma_{i}$ of $K$ sample points, $\left \{ \bm r\left ( {t}_{i}\right ) |i=1,...,K \right \} $, by following this formal procedure:
\begin{equation}
\begin{gathered}
    \hat{C} \left ( \bm r \right ) = \sum_{i=1}^{K} w_{i} \bm c_{i}, \ \ \text{ where }\ \\
     w_{i} = \exp\left ( -\sum_{j=1}^{i-1}\sigma_{j}\delta_{j}\right ) \left ( 1-\exp\left ( -\sigma_{i}\delta_{i} \right )  \right ).
\end{gathered}
\label{eq2}
\end{equation}
Here, $w_{i}$ indicates the weight or hitting probability of $i$-th sampling point~\cite{liu2022neural} and $\delta_i$ is the distance between adjacent samples.

\noindent \textbf{Text-to-3D Scene Editing Method, IN2N,} operates by repeatedly updating the training dataset images using a diffusion model and then training the NeRF on these modified images to maintain a consistent 3D representation. This iterative approach allows the gradual integration of the diffusion priors into the 3D scene, enabling substantial edits. The image-conditioned diffusion model (IP2P~\cite{brooks2023instructpix2pix}) helps preserve the original scene's structure and identity.

\begin{figure}[t]
\centering
\includegraphics[width=0.95\linewidth]{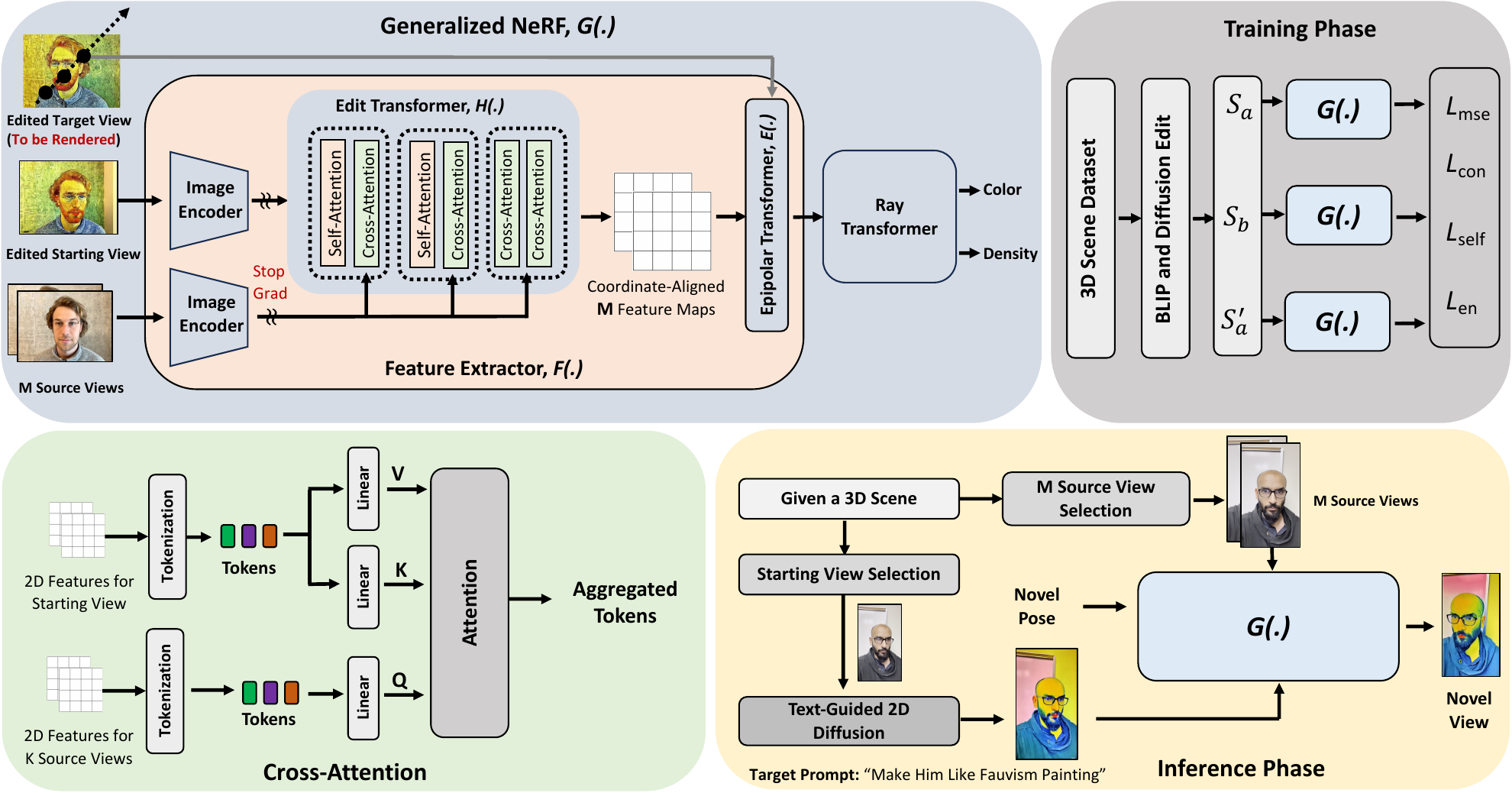}
%\vspace{-1mm}
\caption{\footnotesize \textbf{Overview of our proposed method}. \textbf{\textit{Top Left.}} We train a generalized NeRF ($\mathbf{G}(.)$) model that takes an \emph{edited starting view} and $M$ source views to render a novel target view. Here, the \emph{edited target view} is not the input to the model, rather will be rendered and works as the ground truth for the model output. In $\mathbf{G}(.)$, we employ a novel \emph{Edit transformer} that utilizes: \textbf{\textit{Bottom Left.}} cross-view attention to produce style-informed source feature maps that will be aggregated through an Epipolar transformer. \textbf{\textit{Top Right.}} During training, we employ different sets of source views $S_{a}, S_{b}, S'_{b}$ for 4 different loss functions. Note that $S'_{a}$ is a variant of $S_{a}$ with additional ray information for calculating $\mathcal{L}_{con}$. \textbf{\textit{Bottom Right.}} During inference, only a single image needs to be edited to obtain a 3D-edited scene.} 
\label{fig:framework}
%\vspace{-4mm}
\end{figure}

%\vspace{-3mm}
\subsection{Free-Editor: Zero-shot Scene Editing}
%\vspace{-1mm}
In this section, we describe our proposed method \emph{Free-Editor}, a training-free approach for 3D scene editing without the requirements of \emph{iterative updates of dataset} and \emph{per-scene optimization}. Consider a dataset that contains $L$ number of 3D scenes in terms of images and their camera intrinsic and extrinsic parameters. Let us define a 3D scene training data that contains N images with their corresponding camera parameters, $\{I_l \in \mathbb{R}^{H\times W \times 3}, P_l \in \mathbb{R}^{3 \times 4}\}$. First, we select a \emph{starting view} $I_0$ and render a \emph{target view} $I_{t}$. 
To apply specific 2D editing in $I_{t}$ and $I_0$, we employ a text-guided diffusion model $\mathbf{D}$ with group attention,   
\begin{equation}\label{eq:diff_edit_start}
    \hat{I}_0 = \mathbf{D}(I_0, C^{0}_{in}, C^{0}_{tgt}, \Phi),
    \hat{I}_t = \mathbf{D}(I_t, C^{t}_{in}, C^{0}_{tgt}, \Phi),
\end{equation}
where $\Phi$ is the diffusion model, $C^{0}_{in}$ and $C^{t}_{in}$ are input captions of $I_0$ and $I_t$, respectively. On the other hand, $C^{0}_{tgt}$ is the target caption used for editing. As the next step, we perform \emph{source-view selection} to select $M$ source views, $S = \{I_m, P_m\}_{m=1}^{M}$, from the remaining source views $\{I_n, P_n\}_{n=1}^{N-1}$, where $M<N-1$. Details of selecting $M$ views are in Sec.~\ref{sec:view_sel}. The generalized NeRF model~\cite{johari2022geonerf, liu2022neural} with parameters $\theta$, $\mathbf{G}(.;\theta)$, takes $\{I_m, P_m\}_{m=1}^{M}$ and $\hat{I}_0, P_0$  as inputs and predicts $\tilde{I}_t$ as output,
\begin{equation}
    \tilde{I}_t = \mathbf{G}(\hat{I}_0, I_m, P_0, P_m, \theta | m= 1, \ldots M)
\end{equation}
To train $\mathbf{G}(.;\theta)$, we minimize the following optimization objective, 
\begin{equation}\label{eq:final_eq}
    \argmin_{\theta} \mathcal{L}_{tot}(\theta; \hat{I}_t,\tilde{I}_t).
\end{equation}

\noindent \textbf{Style-aware Multi-view Feature Extraction.}
In Figure~\ref{fig:framework}, we show the details of our proposed method. We can consider the generalized NeRF as a combination of feature extractor ($\bm{F}$) to extract coordinate-aligned feature maps before aggregating them and ray transformer ($\bm{R}$) ~\cite{wang2022attention} that will transform the features into color and density. Using $\bm{F}$), we adopt a feed-forward fashion to extract generalized features from multi-view images $\hat{I}_0, \{I_m, P_m\}_{m=1}^{M}$ and convert them later into 3D representation using $\bm{R}$. Eq.~\ref{eq:nerf} shows how a 3D location can be mapped into color and density. Here, we do that in two stages: i) create a coordinate-aligned feature field using $\bm{F}$), and ii) aggregate point-wise features along different rays to form ray colors using an attention-based ray transformer. We use $\hat{I}_t$ as the ground truth for ray colors. 
 
We first construct hierarchical image features using pre-trained 2D CNN Image Encoder~\cite{he2016deep} as follows, $\bm{f}_i = \bm{T}(I_i); m = 1, \ldots M$, where $\bm{f}_i$ is the image feature for $i^{th}$ source image. We then feed these features to the newly proposed edit transformer, $\bm{h}_i = \bm{H}(\bm{\hat{f}}_0,\bm{f}_i)$, which would give us the editing-informed multi-view feature maps. Here, $\bm{\hat{f}}_0$ is the 2D image feature corresponding to the starting view, $\hat{I}_0$. We use both self-attention and cross-attention in our edit transformer (shown in Fig.~\ref{fig:framework}) with different purposes. The functional mechanism of attention can be defined as $\mathrm{Attention}(Q,K,V)=\mathrm{Softmax}(\frac{Q K^T}{\sqrt{d}}) \cdot V$, where
\begin{equation}
    Q=W^Q z, K=W^K z, V=W^V z.
\end{equation} 
Here, $W^Q$, $W^K$, and $W^V$ denote trainable matrices that project the inputs to the query ($Q$), key ($K$), and value ($V$) components, respectively. $z$ represents the latent features, and $d$ represents the output dimension of the key and query features. The objective of self-attention is to learn long-range and relevant information within a given view. In our particular case, we aim to capture the exact editing effects that have taken place in $\hat{I}_0$ utilizing self-attention.  However, capturing only single-view information using self-attention is not enough as it is required to have multi-view feature maps for a successful novel view synthesis. To understand why multi-view information is needed, we briefly analyze the recently proposed Epipolar Aggregated Transformer~\cite{yang2022mvs2d, suhail2022generalizable} which functions between the target pixels and pixels positioned on the epipolar line of multiple source views. Using the epipolar geometry constraint, the features can be aggregated to capture long-range content information within and across images. However, vanilla multi-view feature maps obtained from $M$ source views do not have the editing information in $\hat{I}_0$. We find a simple fix to this issue by explicitly injecting editing information from $\hat{I}_0$ into the multi-view source feature maps with the help of cross-attention. As shown in Fig.~\ref{fig:framework}, we first tokenize the 2D image features obtained from $\bm{T}$ which reduces the attention complexity significantly. To extract the key ($K$) and value ($V$), tokens from the starting view are used whereas we use source view tokens as the query ($Q$). Using cross-attention, we aggregate features from source views towards $\hat{I}_0$.  

Now, for each target pixel in $\hat{I}_t$, we uniformly sample $P$ coordinate-aligned 3D points $\{\bm{x}_1, \ldots, \bm{x}_P\}$ from the set of points between far and near planes. Each of these points is projected into the feature maps obtained from $\bm{H}$ and then aggregated to form a coordinate-aligned feature field as follows,
\begin{equation}\label{eq:coardinate_feat}
    \bm{F}(x_p, \phi) = \bm{E}(\bm{h}_1(\Pi_1(\bm{x}_p))), \ldots, \bm{h}_M(\Pi_M(\bm{x}_p))).
\end{equation}
Here, $\bm{E}$ is the Epipolar Aggregated Transformer, $\Pi_i(\bm{x}_p)$ projects 3D point $\bm{x}_p$ onto the $i$-$th$ source-image plane with the help of an extrinsic matrix. We use bilinear interpolation to compute the feature vector $\bm{h}_1({\bm{u}})$ at projected 2D position $\bm{u} \in \mathbb{R}^2$. Finally, we utilize the Ray Transformer $\bm{R}$ along with an MLP to dynamically learn the blending weights along the ray for each point and produce the RGB color information. Given ray information, $\bm{r}$ and $\{\bm{x}_1, \ldots, \bm{x}_P\}$ as the uniformly sampled points along $\bm{r}$, an MLP ($\bm{V}$) can be used to map the pooled features vectors from $\bm{R}$ to RGB color $\tilde{\bm{C}}_{t}$,
\begin{equation}
    \tilde{\bm{C}}_{t}(\bm{r}) = \bm{V}(\bm{R}(\bm{F}(\bm{x}_p, \phi))); \hspace{1mm} p \in [1, P]. 
\end{equation}

For training the model end-to-end, we employ different loss functions that serve important roles in producing good performance. 

% \medskip

%\vspace{-3mm}
\subsection{Training Objectives and Data Generation}\label{sec:loss_functions}
%\vspace{-1mm}

\noindent \textbf{Photometric Loss, $\mathcal{L}_{mse}$.} First, we adopt the photometric loss in NeRF~\cite{mildenhall2021nerf} which is defined as the mean square error (MSE) between the predicted and ground truth pixel colors,
\begin{equation}\label{eq:mse}
    \mathcal{L}_{mse} = \sum_{r \in \mathcal{R}} ||\tilde{\bm{C}}_{t}(\bm{r}) - \hat{\bm{C}}_{t}(\bm{r})||^2,
\end{equation}
where $\mathcal{R}$ represents the set of rays and $\hat{\bm{C}}_{t}(\bm{r})$ is the ground truth pixel values in $\hat{I}_t$ for ray $\bm{r} \in \mathcal{R}$.

\noindent \textbf{Multi-view Consistency Loss, $\mathcal{L}_{con}$.} As we are editing only a single image to edit an entire 3D scene, achieving spatial smoothness is challenging due to the constraints of view geometry in NeRF. To tackle this issue, we introduce a multi-view consistency loss to encourage a smooth transition between texture or color between neighboring views. Let us denote the feature distribution obtained at 3D point $\bm{x}^j_p$ as $\bm{e}^{j}_p = \{\bm{h}_1(\Pi_1(\bm{x}^j_p))), \ldots, \bm{h}_M(\Pi_M(\bm{x}^{j}_p))\}$, obtained from $M$ source views. Here, the point $\bm{x}^j_p$ is sampled along the ray, $\bm{r}_j$. Let us select another ray $\bm{r}_{j'}$ which is very close to $\bm{r}_j$. For each point $\bm{x}^j_p$, we select its closest point $\bm{x}^{j'}_p$ along the ray $\bm{r}_{j'}$ based on their Euclidean distance, denoted as $d^{p}_{j,j'} = ||\bm{x}^j_p - \bm{x}^{j'}_p||$. To encourage consistency among the coordinate-aligned features of the closest points, we employ Jensen-Shannon Divergence (JSD) loss,
\begin{equation}
    \mathcal{L}_{J}(\bm{x}_p) = JSD(\bm{e}^{j}_p||\bm{e}^{j'}_p),
\end{equation}
where $\bm{e}^{j'}_p$ is the features corresponding to $\bm{r}_{j'}$. We use JSD loss for its symmetric nature which offers some notable advantages over Kullback–Leibler (KL) divergence~\cite{huszar2015not} loss. Since closer points in the pixel space should have smaller distances in the feature space, we employ a weighted JSD loss for defining our final multi-view consistency loss, 
\begin{equation}\label{eq:con}
    \mathcal{L}_{con} = \sum_{p=1}^{P} \omega_p \mathcal{L}_{J}(\bm{x}_p).
\end{equation}
Here, $\omega_p$ indicates the weight corresponding to the pair $(\bm{x}^j_p,\bm{x}^{j'}_p)$ which can be expressed as $\omega_p = \frac{e^{-d^{p}_{j,j'}}}{\sum_{p=1}^{P} e^{-d^{p}_{j,j'}}}$. Our unique formulation of $\mathcal{L}_{con}$ imposes consistency on 3D points across various viewpoints, inherently promoting smoothness in the scene's geometry. 
\begin{figure}[t]
    \centering
    \includegraphics[width=0.9\linewidth]{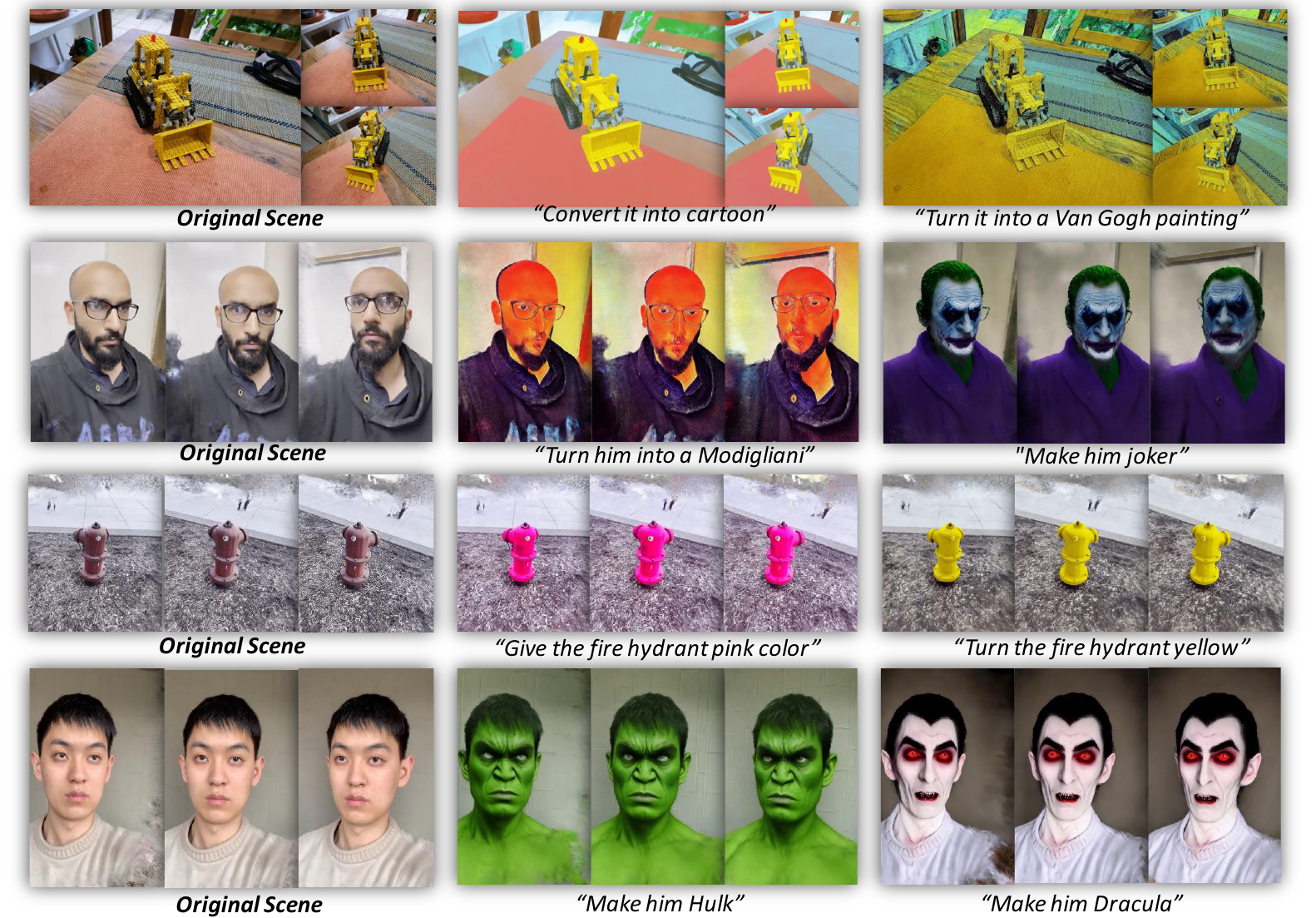}
    %\vspace{-3mm}
    \caption{\footnotesize \textbf{Text-driven 3D scene editing.} Illustration of text-driven 3D scene editing using our proposed method across various target poses. This figure showcases the view-consistent results generated by our method. A qualitative evaluation on multiple scenes reveals the efficacy of our approach: starting from a single view, our method successfully generates novel views that are conditioned on the editing prompt, demonstrating its robustness and versatility in 3D scene editing. }
    \label{fig:text_to_3D}
    %\vspace{-4mm}
\end{figure}

\noindent \textbf{Self-View Robust Loss, $\mathcal{L}_{self}$.} In general, when the training data for a 3D scene remains coherent, generating the same target view using different source views usually produces consistent outcomes. However, this may not hold true for our case as we are dealing with an edited target view. To address this, we choose two different sets of source views $S_a = \{I^a_m, P^a_m\}_{m=1}^{M}$ and $S_b = \{I^b_m, P^b_m\}_{m=1}^{M}$. The predicted target views utilizing $S_a$ and $S_b$ are $\hat{I}^a_t$ and $\hat{I}^b_t$, respectively which should have consistent content information. To ensure this consistency, we employ 
\begin{equation}\label{eq:self}
    \mathcal{L}_{self} = \sum_{r \in \mathcal{R}} ||\tilde{\bm{C}}^a_{t}(\bm{r}) - \tilde{\bm{C}}^b_{t}(\bm{r})||^2,
\end{equation}
where $\tilde{\bm{C}}^a_{t}$ and $\tilde{\bm{C}}^b_{t}$ indicate RGB color values in $\hat{I}^a_t$ and $\hat{I}^b_t$, respectively.

% \medskip

\noindent \textbf{Entropy Loss, $\mathcal{L}_{en}$.} In addition, we consider an entropy loss for regularizing the hitting probabilities of the sampled points~\cite{kim2022infonerf}, 
\begin{equation}\label{eq:entropy}
    \mathcal{L}_{en} = - \sum w_i \log(1-w_i).
\end{equation}

Finally, the total loss function employed to train our framework can be expressed as follows, 
\begin{equation}\label{eq:total}
     \mathcal{L}_{tot} = \mathcal{L}_{mse} + \lambda_c\mathcal{L}_{con} +  \lambda_s\mathcal{L}_{self} + \lambda_e\mathcal{L}_{en},
\end{equation}
where $\lambda_c$, $\lambda_s$, are $\lambda_e$ the loss coefficients.

\noindent \textbf{Training Data Generation.}\label{sec:view_sel}
From each scene, we first select a target view and then identify a pool of $m.(M+1)$ nearby views ($m$ is sampled
uniformly at random from [1, 3]), among which a randomly sampled view is chosen as the \emph{starting view} while $M$ other views are the source views. This sampling approach of $m$ mimics diverse view densities during the training process, enhancing the network's ability to generalize across different view densities. We get the RGB images ${I}_0$ and ${I}_t$ corresponding to the starting and target views, respectively. For editing ${I}_0$ and ${I}_t$, we utilize the open-source pre-trained models BLIP~\cite{li2023blip} and IP2P~\cite{brooks2023instructpix2pix}. The BLIP model produces the input caption $C^0_{in}$ of the \emph{starting view} ${I}_0$. Later, we employ a GPT model to generate $C^0_{tgt}$ by modifying $C^0_{in}$. In addition to GPT, we apply manually designed prompts as well. For example, $C^0_{tgt}$ can be generated by simply following this format- \emph{"\textcolor{cyan}{\textbf{X}} painting of $C^0_{in}$"}. \emph{\textcolor{cyan}{\textbf{X}}} can be chosen from ["Leonardo da Vinci", "Sam Francis", "Max Ernst", "Henri Matisse", "Eva Hesse", "Carl Andre", "Cy Twombly"]. Finally, we feed $C^0_{tgt}$ and ${I}_0$ to IP2P to produce $\hat{I}_0$. For $\hat{I}_t$, we generate multiple edited copies and select the one as the ground truth (for G) that has the highest CLIP consistency score with $\hat{I}_0$. In our work, we use $e_n$ randomly chosen $C^0_{tgt}$ for each scene where $e_n$ is set to be 6. We do not choose a higher value for $e_n$ as our objective is not to learn all types of editing available but rather how to transfer the edits from the \emph{starting view} to other views. During training, we randomly select $M$ from a uniform distribution of [8, 12]. More on this in the \emph{\textit{\textcolor{blue}{supplementary}}}.  
\begin{table}[t]
\begin{minipage}{0.535\textwidth}
    \centering
    \caption{\footnotesize \textbf{Quantitative assessment of 3D scene editing focusing on text alignment and frame consistency} is conducted.  Our method exceeds all other state-of-the-art editing techniques in terms of Edit PSNR metric while achieving comparable performances in other scenarios. We use LLFF dataset here.}
    %\vspace{-2mm}
    \resizebox{1\columnwidth}{!}{
    \begin{tabular}{l|cccccc}
    \toprule
     \textbf{Metrics}              & \textbf{C-NeRF}   & \textbf{NeRF-Art}  & \textbf{IN2N}  & \textbf{DreamEditor}   & \textbf{Ours}   \\
    \midrule
    {Edit PSNR} & 22.15 & 20.89 & 22.26 & 22.34 & \textbf{22.47} \\
    {CTDS} & 0.2375 & 0.2503 & \textbf{0.2804} & 0.2788 & 0.2601 \\
    % {NIQE} & 22.15 & 20.89 & 21.76 & 25.34 & 28.47 \\
    {CDC} & 0.9672 & 0.9751 & \textbf{0.9882} & 0.9850 & 0.9781 \\
    \bottomrule
    \end{tabular}
    }
    \label{tab:quant_comp}
\end{minipage}
\hfill
\begin{minipage}{0.405\textwidth}
    \centering
   \caption{\footnotesize \textbf{PSNR comparison with recent SOTA generalized NeRF methods}. Here, \emph{LLFF-E} indicates the performance on the edited LLFF dataset.}
    % \setlength\tabcolsep{3pt}
    %\vspace{-1.5mm}
    \resizebox{0.8\linewidth}{!}{
     \begin{tabular}{l|cc}
    \toprule
    \textbf{Method}   & \textbf{LLFF}  & \textbf{LLFF-E} \\
    \midrule
    PixelNeRF & 18.66 & 11.03 \\
    MVSNeRF & 21.18 & 16.74 \\
    IBRNet & 25.17 & 19.05 \\
    Neuray & 25.35 & 18.31 \\
    GeoNeRF & \textbf{25.44} & 18.98 \\
    \textsc{Free-Editor} (Ours)  & 24.61 & \textbf{22.47} \\
    \bottomrule
    \end{tabular}%
   }
   \label{tab:generalization_per}
\end{minipage}
    % %\vspace{-5mm}
\end{table}

\noindent \textbf{Inference Phase.}  During inference, we use different sets of scenes and target captions ($C^0_{tgt}$) than the training stage. This is to ensure that \emph{Free-Editor} is generalizable in terms of both scenes and editing prompts.  To edit a test scene, we first randomly select ${I}_0$ that will be edited and $M (= 12)$ source images. We then pass a novel target pose (close to the $M$ source views, but not necessarily close to the ${I}_0$) to render an edited target view. As \emph{Free-Editor} can consistently transfer the edits from one \emph{starting view} to all other views, we can easily edit a 3D scene by editing a sufficient number (e.g. 70–80) of target views. Note that our proposed method edits any particular target view only once, in contrast to the iterative edits proposed in IN2N.

\section{Experiments and Analysis}\label{sec:Experimental_results}
% %\vspace{-1.5mm}
\noindent \textbf{Datasets.} Our model is trained on datasets including Google Scanned Objects~\cite{downs2022google}, NerfStudio~\cite{tancik2023nerfstudio}, Spaces~\cite{flynn2019deepview}, and IBRNet-collect~\cite{wang2021ibrnet}, and 
 \begin{wrapfigure}{r}{0.55\textwidth}
\centering
% %\vspace{-5mm}
    \includegraphics[width=0.9\linewidth]{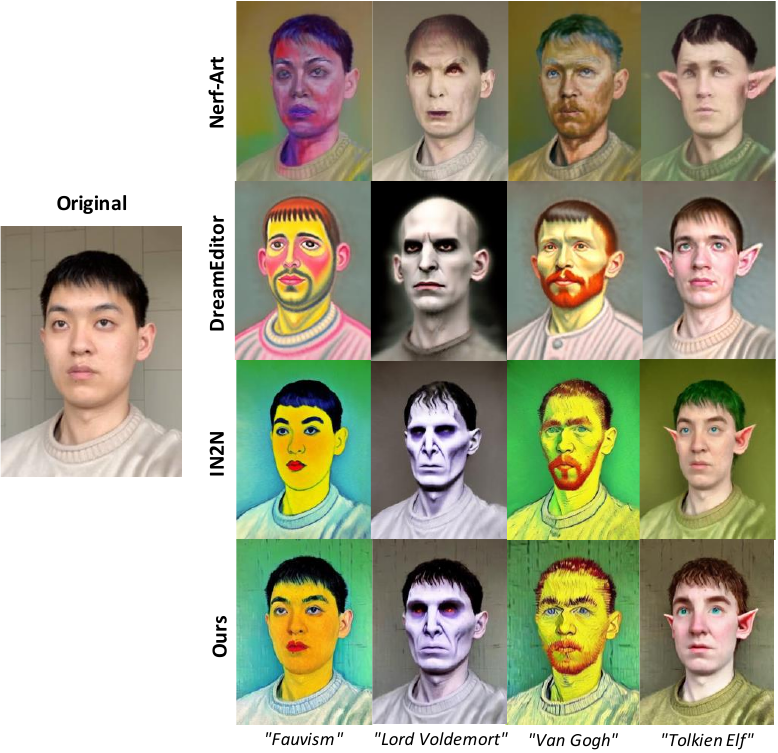}
    %\vspace{-3mm}
    \caption{\footnotesize\textbf{Style Transfer Comparison.} Exhibiting proficiency in conducting style edits within 3D NeRF Scenes, our method exemplifies its versatility and precision through intricate modifications and advanced prompt-guided editing in a three-dimensional environment. Visually, our outcomes resemble those of IN2N, since both methods utilize IP2P for 2D image editing. However, our method tends to preserve background details more effectively than IN2N.}
\label{fig:style_comparison}
% %\vspace{-4.5mm}
\end{wrapfigure}RealEstate10K~\cite{zhou2018stereo}. For evaluation, we use IN2N~\cite{haque2023instruct}, NeRF-Synthetic~\cite{Mildenhall20}, LLFF~\cite{mildenhall2019local}, and our own dataset of four scenes. 

% \medskip

% We use an Adam optimizer with learning rates of 2e-4 for the Edit Transformer, 1e-4 for the Epipolar transformer, and 5e-4 for the Ray Transformer. In addition, a batch size of 2048 rays has been employed. In every batch, we select rays from eight distinct scenes. Within each scene, we randomly sample rays from a singular image for training purposes. As for the loss coefficients, we consider 5e-4, 1e-3 and 1e-4 for  $\lambda_c,  \lambda_s, \text{and} \hspace{1mm} \lambda_e$, respectively.

\noindent \textbf{Training Details.}  is in \emph{\textit{\textcolor{blue}{supplementary}}}.

% \medskip

\noindent \textbf{Baselines.} We report qualitative and quantitative comparisons against four baseline NeRF editing methods including IN2N \cite{haque2023instruct}, NeRF-Art \cite{wang2023nerf}, C-NeRF \cite{wang2022clip}, and DreamEditor \cite{zhuang2023dreameditor}.
The default 2D-image editing model is Instruct-Pix2Pix (IP2P)~\cite{brooks2023instructpix2pix}
%\vspace{-3mm}
\subsection{Qualitative Results}
%\vspace{-1mm}
Figures~\ref{fig:text_to_3D}, \ref{fig:style_comparison} demonstrate the proficiency of our method in executing effective style edits, maintaining 3D coherence, and conforming to textual narratives.

 % \textbf{PSNR}  & 21.64 & 21.98  & 22.29 & 22.38 &  22.44 & 22.47 & \textbf{22.52}   \\
 %     \textbf{CDC}  & 0.963  & 0.965 & 0.972 & 0.975 & 0.977 & \textbf{0.978} & 0.978 \\

\noindent \textbf{Text-driven 3D Scene Editing.} Figure~\ref{fig:text_to_3D} shows the text-driven editing performance of our proposed method. We use different text-scene pairs to show the diversity of our method. We start with the edited starting view and produce the novel target view from different poses.  Our method shows notable alignment between the provided text description and the resulting views. Yet, there are instances where the alignment isn't perfect. For instance, when attempting to transform an image into Modigliani style, the details around the nose might not be accurately captured. This kind of intricate detail can be challenging to represent when working within significant limitations during the editing process. 
% More on this is in \emph{\textit{\textcolor{blue}{supplementary}}}.
% Compared to other text-driven editing methods (Figure~\ref{fig:text-methods-com}), our \modelname{} directly infers editing results from the given text, avoiding the retraining required by NeRF-Art and Instruct-NeRF2NeRF. 

\noindent \textbf{Style Transfer.} Figure~\ref{fig:style_comparison} illustrates the visual comparisons between \textsc{Free-Editor} and other 3D scene style transfer methods.  Owing to the use of the IP2P~\cite{brooks2023instructpix2pix} editing backbone, our editing outcomes closely resemble those of IN2N~\cite{haque2023instruct}. However, the distinct advantage of our approach lies in its training-free nature, setting it apart from others. A notable distinction is that our method tends to preserve background details more effectively than IN2N~\cite{haque2023instruct}, which often struggles to maintain this aspect. Comparisons with top-tier methods further corroborate the effectiveness of our technique and show the adeptness of our method in capturing both the color palette and stroke patterns of the desired style. Furthermore, our results are more realistic and preserve non-targeted areas, facilitating multiple sequential edits. Details are in \emph{\textit{\textcolor{blue}{supplementary}}}.

% figure}[t]
% %\begin{center}

% \end{figure}

% Additionally, \textsc{Free-Editor} achieves editing results comparable to those of IN2N, but without the requirement for re-training the model.
% \paragraph{Appearance Editing}: Demonstrated in Figure~\ref{fig:appearance}, \textsc{Free-Editor} accurately identifies and edits target areas consistent with descriptions, outperforming models like NeRF-Art~\cite{wang2023nerf} and IN2N~\cite{haque2023instruct} in efficiency and accuracy.

\begin{table}[t]
\begin{minipage}{0.45\linewidth}
  \centering
     \caption{\footnotesize \textbf{An ablation with the number of source views, M.} A higher value of M produces slightly better performance before the performance saturates at a certain point. LLFF dataset has been used.}
     %\vspace{-2mm}
  \setlength\tabcolsep{3pt}
  \resizebox{0.62\linewidth}{!}{
     \begin{tabular}{l|ccc}
    \toprule
    \textbf{M} & \textbf{Edit PSNR}  & \textbf{CTDS} & \textbf{CDC} \\
    \midrule
    3 & 20.94 & 0.2386 & 0.953\\
    4 & 21.68 & 0.2471 & 0.962 \\
    6 & 22.09 & 0.2548 & 0.972\\
    8 & 22.38 & 0.2563 & 0.975 \\
    10 & 22.44 & 0.2590 & 0.977 \\
    12 & 22.47 & 0.2601 & \textbf{0.978}\\
    18 & \textbf{22.52} & \textbf{0.2604} & 0.978  \\
    \bottomrule
    \end{tabular}
   }
   % %\vspace{-2mm}

   \label{tab:source_views_ablation}
\end{minipage}
\hspace{0.02\columnwidth}
\begin{minipage}{0.45\linewidth}
  \centering
  \caption{\footnotesize \textbf{Quantitative ablation study with different loss functions}. Self-view robust loss impacts the color consistency while $\textit{L}_{con}$ impacts the smooth transfer of color information. Our own scenes have been used for this study.}
  %\vspace{-2mm}
  \setlength\tabcolsep{3pt}
  \resizebox{0.7\linewidth}{!}{
    \begin{tabular}{c|cccc}
    \toprule
    Use Cases &   w/o $\textit{L}_{con}$ & w/o $\textit{L}_{self}$  & All Loss \\
    \midrule
    Case 1  & 20.98 & 19.63  & \textbf{22.76} \\
    Case 2  & 22.86  & 21.52  & \textbf{23.11} \\
    Case 3  & 23.14 & 22.29  & \textbf{24.21} \\
    Case 4  & 23.08 & 21.81  & \textbf{23.96} \\
    Case 5  & 21.37 & 20.13  & \textbf{22.56} \\
    Case 6  & 21.25 & 20.38  & \textbf{22.03} \\
    \bottomrule
    \end{tabular}%
   }
% %\vspace{-2mm}

\label{tab:ablation_study}   
\end{minipage}

%\vspace{-5mm}
\end{table}

%\vspace{-3.5mm}
\subsection{Quantitative Results}
%\vspace{-1.5mm}
\noindent \textbf{3D Scene Editing.}
The chosen metrics for the editing evaluation are CLIP Text-Image Directional Similarity (CTDS) and CLIP directional consistency (CDS), which serve as indicators of how effectively each method preserves 3D consistency across edited scenes. CTDS evaluates how well the executed 3D edits correspond to the text instructions. While CDS is akin to CTDS it assesses the similarity in direction between the original and edited images in successive frames along newly generated camera paths. Additionally, \emph{Edit PSNR} compares the cosine similarity and PSNR between each rendered view from the edited and input NeRF. These metrics collectively provide insight into the integrity of the 3D scene post-editing. Table~\ref{tab:quant_comp} showcases quantitative evaluations on the LLFF dataset across diverse scene editing tasks. Our findings indicate that both IN2N and our proposed approach produce outcomes consistent with the original viewpoints, demonstrated by their CLIP directional scores. Notably, the proposed method outperforms IN2N in terms of Edit PSNR, signifying better preservation of scene consistency with the original input. This suggests that our method maintains the details of the initial scene while implementing edits more effectively. Despite these advantages, our method slightly lags behind IN2N in overall performance, likely due to the zero-shot nature of \textsc{Free-Editor}. However, our findings underscore the strength of the proposed method in preserving scene integrity during editing.

\noindent \textbf{Generalization Capabilities.} Although the goal of our work is scene editing, we employed a generalized NeRF to achieve zero-shot capabilities. Therefore, it necessitates us to validate our method in generalization tasks too. For this, we consider recent generalized NeRF methods such as pixelNeRF~\cite{yu2021pixelnerf}, IBRNet~\cite{wang2021ibrnet}, MVSNeRF~\cite{chen2021mvsnerf}, Neuray~\cite{liu2022neural}, GeoNeRF~\cite{johari2022geonerf}. We modify the original forward-facing LLFF data dataset, LLFF-E based on diffusion-based editing, e.g. color or attribute changes.
Table~\ref{tab:generalization_per} shows our findings for this particular experiment. For the regular LLFF dataset, \textsc{Free-Editor} obtains slightly worse performance than previous methods which is somewhat understandable as it is developed mostly for scene editing. However, these methods severely underperform when we evaluate them on LLFF-E. This shows that one can not just use an off-the-shelf generalized NeRF for editing purposes. It also shows the necessity of developing a proper technique to obtain a generalizable method with editing capabilities. More on this is in \emph{\textit{\textcolor{blue}{supplementary}}} 

%\vspace{-4mm}
\subsection{Ablation Studies}
%\vspace{-2mm}
In this section, we perform an ablation study with different loss functions and the efficiency of \textsc{Free-Editor}.

\begin{figure}[t]
\begin{minipage}{0.455\textwidth}
\centering
%\vspace{-2.5mm}
    \includegraphics[width=0.95\linewidth]{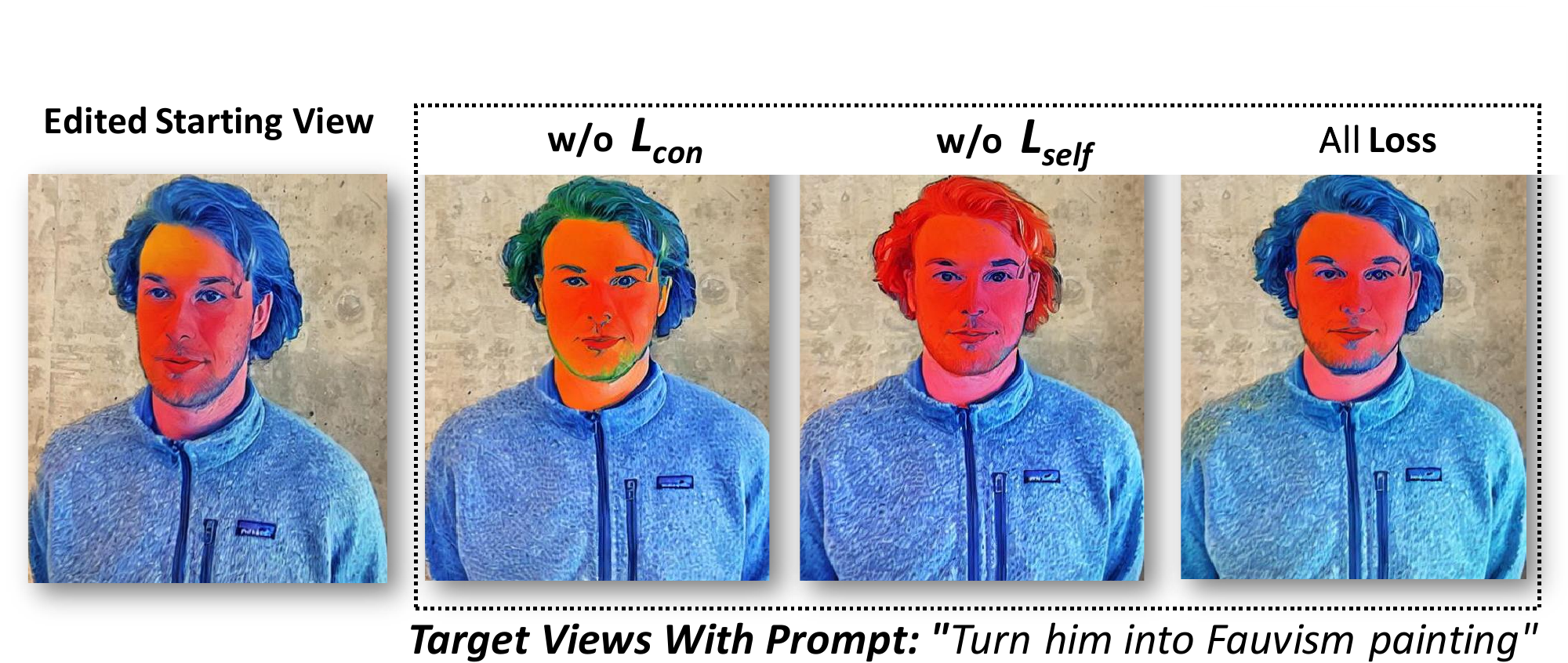}
%    \end{center}
    %\vspace{-2mm}
    \caption{\footnotesize \textbf{Loss Sensitivity.} An ablation study, to show the impact of different loss functions on the final performance.}
\label{fig:ablation}
\end{minipage}
\hfill
\begin{minipage}{0.485\textwidth}
    \centering
    \includegraphics[width=1\linewidth]{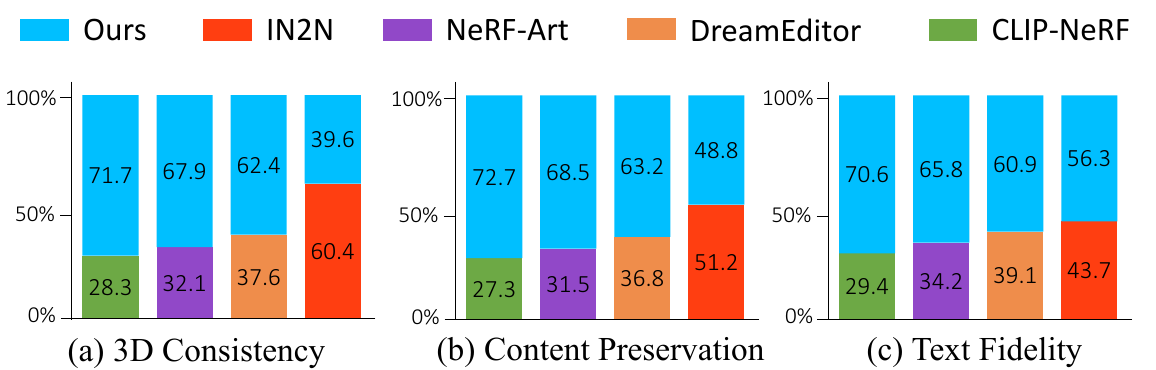}
    \caption{ \footnotesize In an extensive \textbf{user study} that assessed three evaluation metrics, our approach demonstrated comparable performance to IN2N~\cite{haque2023instruct}}
\label{fig:user_study}
\end{minipage}
    % %\vspace{-5mm}
\end{figure}

\noindent \textbf{Effect of Different Loss Functions.} We study the impact of different\begin{wraptable}{r}{0.55\textwidth}
%\vspace{-8.5mm}
% Our proposed method achieves significantly faster editing time and better space complexity.
\centering
\caption{\footnotesize \textbf{Runtime efficiency comparison of different methods}. We take 2 of our scenes and apply 10 different editing before averaging. }
  \resizebox{1\linewidth}{!}{
    \begin{tabular}{c|cccc}
    \toprule
    Method & PSNR (dB) & Edit-time (mins.) & Time Complexity  & Space Complexity \\
    \midrule
        Clip-NeRF & 21.15 & 1034.2 & O(n)   & O(n)\\
        NeRF-Art  & 21.64 & 780.6  & O(n)  & O(n)\\
        Instruct-N2N & 22.98 & 62.1  & O(n)  & O(n) \\
        DreamEditor & \textbf{23.18} & 70.5 & O(n)  & O(n)\\
        \textbf{Ours} & 23.06 & \textbf{3.2} & O(n) & \textbf{O(1)}\\
    \bottomrule
    \end{tabular}}
%\vspace{-7mm}
\label{tab:editing_time}   
\end{wraptable} loss functions on the overall performance of our proposed method. Table~\ref{tab:ablation_study} shows 6 different use cases (scenes) where we apply 4 different text prompts for each scene. It can be observed that $\textit{L}_{self}$ has the most impact on the editing performance.  The reason behind this is the compromised generalizability of the model without $\textit{L}_{self}$. On the other hand, $\textit{L}_{con}$ helps us obtain better spatial smoothness, which can be also shown by the qualitative comparison we present in Fig.~\ref{fig:ablation}. Note that the impact of $\textit{L}_{con}$ wears off when we increase $M$. 

\noindent \textbf{Model Efficiency.} One of the main goals is to edit a particular 3D scene within a realistic timeframe. Table~\ref{tab:editing_time} shows that we are close to achieving that goal. \textsc{Free-Editor} obtains almost  20$\times$ better runtime efficiency compared to the previous SOTA. Our proposed approach reduces the total editing time while obtaining better space efficiency, leading to a constant space complexity of ${O}(1)$. On the contrary, previous methods necessitate the retraining of a model for each distinct scene or editing type, resulting in increased space complexity $O(n)$. 

\noindent \textbf{Effect of M.} In Table~\ref{tab:source_views_ablation}, we study the impact of different numbers of source views. It can be observed that our proposed method can produce similar performance even with very few source views. The performance trade-off is not meaningful for $M>12$.

%\vspace{-3mm}
\subsection{User Study}
%\vspace{-1mm}
We conducted a user study to observe the acceptability of our method in comparison to other leading-edge methods. This study involved a broad participant base, resulting in a total of 1000 responses across three critical evaluation criteria: the 3D spatial coherence, the retention of the original scene's elements, and the accuracy in reflecting the given textual descriptions. The outcomes of this user survey are visually represented in Figure~\ref{fig:user_study}. These results indicate a preference for results generated by our method and IN2N~\cite{haque2023instruct}, highlighting Free-Editor's proficiency in these key areas. Detailed information regarding the methodology and execution of this user study is provided in the \emph{\textit{\textcolor{blue}{supplementary}}}.

% \begin{wrapfigure}{r}{0.55\textwidth}
% \centering
% %\vspace{-2.5mm}
%     \includegraphics[width=0.95\linewidth]{sec/Fig/ablation_3.pdf}
% %    \end{center}
%     %\vspace{-2mm}
%     \caption{\footnotesize \textbf{Loss Sensitivity.} An ablation study, to show the impact of different loss functions on the final performance.}
% \label{fig:ablation}
% %\vspace{-7mm}
% \end{wrapfigure} 

% \begin{wrapfigure}{r}{0.5\textwidth}
% %\vspace{-7mm}
%     \includegraphics[width=1\linewidth]{sec/Fig/nazmul_user_study.pdf}
% \caption{ \footnotesize In an extensive \textbf{user study} that assessed three evaluation metrics, our approach demonstrated comparable performance to IN2N~\cite{haque2023instruct}}
% \label{fig:user_study}
% %\vspace{-8mm}
% \end{wrapfigure} 
%\vspace{-3mm}
\section{Discussion and Limitations}
%\vspace{-2mm}
One potential solution to the issue of \emph{multi-view inconsistency} within the same scene can be through \emph{trial and error}. Specifically, we can generate a particular set of edited images and then observe the rendering performance. Since we are using a pre-trained 2D diffusion model, this process can be repeated until we get our desired editing effects in the target view. However, it may take hundreds of \emph{trial and error} iterations before achieving a reasonable performance on the edited scene. Therefore, developing an efficient method for 3D scene editing is necessary and makes practical sense. 
% Another way we can tackle the problem of \emph{multi-view inconsistency} is to design a view-filtering system that will give us the set of edited source images with the best consistency score. For calculating this score, we can use CLIP~\cite{radford2021learning} score as well as the model feedback. A good design of the filtering system can reduce the 3D inconsistency among the edited images significantly.   

\subsubsection{Limitations.}  Since we depend on the 2D image pre-editing process~\cite{brooks2023instructpix2pix} for such edits, \emph{multi-view inconsistency} could still be an issue here. To tackle this, we can use the CLIP consistency score to see whether we have a good match between the starting and target views. However, the probability of inconsistent edits between 2 views is much lower as compared to the scenario where we need to edit all training images. Another limitations of our work is to heavily focusing on style transfer as there are other area of interests such as object addition or removal. We leave this to the future studies of our work. Another limitation could be that for complex and large scenes, the model may need fine-tuning, not full training, for the desired performance. 

% Failures of editing this single view adversely affect 3D scene editing outcomes. This is a crucial factor in our work as the success of our method heavily relies on the success of 2D editing.  For this, 

% \subsubsection{Why do we need FreeEditor?} Bring from the rebuttal...

% \textcolor{red}{Shorten or rewrite above limitations.... Bring some of the limitations from CVPR rebuttal...............}
%\vspace{-3mm}
\section{Conclusion}
%\vspace{-2mm}
We proposed a zero-shot text-driven 3D scene editing technique that does not require any re-training. Although the issue of re-training can be addressed by training a generalized NeRF model, the produced features do not contain the necessary editing information. To overcome this, our proposed edit transformer can effectively transfer the style information to the rendered target views through cross-attention. In addition, multi-view consistency loss and self-view robust loss are employed to further enhance spatial smoothness and color consistency. Our method offers not only diverse editing capabilities but also considerable benefits in processing speed and storage efficiency when compared with prior methods with requirements of retraining for individual scenes or modifications.

% \section{Conclusion}
% The paper ends with a conclusion and. 
% \clearpage\mbox{}Page \thepage\ of the manuscript.
% \clearpage\mbox{}Page \thepage\ of the manuscript.
% \clearpage\mbox{}Page \thepage\ of the manuscript.
% \clearpage\mbox{}Page \thepage\ of the manuscript.
% \clearpage\mbox{}Page \thepage\ of the manuscript. This is the last page.
% \par\vfill\par
% Now we have reached the maximum length of an ECCV \ECCVyear{} submission (excluding references).
% References should start immediately after the main text, but can continue past p.\ 14 if needed.
% \clearpage  % TODO REVIEW/FINAL: This \clearpage needs to be removed from both review and camera-ready versions.

% ---- Bibliography ----
%
% BibTeX users should specify bibliography style 'splncs04'.
% References will then be sorted and formatted in the correct style.
%
\bibliographystyle{splncs04}
\bibliography{egbib}
\end{document}